# Deep Multi-Species Embedding

Di Chen
chendi9412@sjtu.edu.cn
Shanghai JiaoTong University

Yexiang Xue
yexiang@cs.cornell.edu
Cornell University

Daniel Fink
df36@cornell.edu
Cornell Lab of Ornithology

Shuo Chen
shuochen@cs.cornell.edu
Cornell University

Carla Gomes
gomes@cs.cornell.edu
Cornell University

February 21, 2017

## Abstract

Understanding how species are distributed across landscapes over time is a fundamental question in biodiversity research. Unfortunately, most species distribution models only target a single species at a time, despite strong ecological evidence that species are not independently distributed. We propose Deep Multi-Species Embedding (DMSE), which jointly embeds vectors corresponding to multiple species as well as vectors representing environmental covariates into a common high-dimensional feature space via a deep neural network. Applied to bird observational data from the citizen science project *eBird*, we demonstrate how the DMSE model discovers inter-species relationships to outperform single-species distribution models (random forests and SVMs) as well as competing multi-label models. Additionally, we demonstrate the benefit of using a deep neural network to extract features within the embedding and show how they improve the predictive performance of species distribution modelling. An important domain contribution of the DMSE model is the ability to discover and describe species interactions while simultaneously learning the shared habitat preferences among species. As an additional contribution, we provide a graphical embedding of hundreds of bird species in the Northeast US.

## 1 Introduction

Understanding the spatial distribution of species and how species interact with each other and their environment is essential for developing science-based conservation plans and ecological research. However, most species distribution models only target a single species at a time [25, 7, 8]. These single-species models ignore the role of species interactions like competition for shared resources (food, territory, etc.). For example, American Robin and Blue Jay are likely to be seen in the same place since the Blue Jay preys on Robin's eggs or fledglings and sometimes even steals its nest. Therefore, a model that predicts the occupancy of a collection of species instead of modelling each species individually is needed. The most straightforward formulation of a multi-species model [31] directly considers the probability of seeing a collection of species. However, this direct approach suffers from combinatorial intractability due to the large number of possible ways to form the collection. As a result, an efficient method of jointly modelling species distribution in a large scale is still lacking.

We propose a novel method called Deep Multi-Species Embedding which can jointly model the distribution of hundreds of species as well as the correlation among species. DMSE jointly embeds multiple species as well as environmental covariates into a high-dimensional feature vector space via a deep neural network. Each embedded vector carries semantic meaning to the modeled entity, and the inner products between them capture the relationships between entities (such as environmental preference or correlation between species).

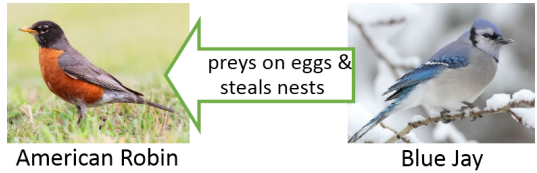

Figure 1: The ecological relationship between American Robin and Blue Jay.

Applied to *eBird* bird observational data [24], we demonstrate how the DMSE model discovers inter-species relationships to outperform the predictions of single-species distribution models (random forests and SVMs) as well as competing multi-label models. As the number of species goes up, the gap in predictive performance between our multi-species DMSE and baseline models keeps widening. Additionally, we demonstrate the benefit of using a deep neural network for feature extraction and show how the features improve the quality of species distribution modelling.

We also show a visualization of the embedding for hundreds of bird species in the Northeast US. It provides an intuitive picture about species environmental preferences and the correlations among each other. Through this model, we are also able to quantitatively measure many species-species interaction which could only be qualitatively described by ecologists before.

## 2 Multi-Species Modelling

Our goal is to estimate the joint distribution of multiple species based on the observational data recording the presence or absence of each species at a site. More formally, given a collection of species $\{species_1, ..., species_n\}$ and the species observation data $D = \{(b_1, l_1), ..., (b_N, l_N)\}$, we would like to estimate the distribution $\Pr(b_i|l_i)$. Here $b_i \in \{0,1\}^n$ is an indicator for the species co-occurrence of each observation, $b_{i,j} = 1$ if and only if $species_j$ was detected at site $i$, and $l_i = (f_1, ..., f_m)^T$ is an environmental feature vector that contains the values of $m$ environmental covariates (or features) that describe site $i$. To simplify notation, we also use $l_i$ to denote the observation site $i$.

### 2.1 From One Species to More

Our DMSE method is based on the latent variable formulation of the probit model [4] which is widely used to model binary outcomes. For the clarity of presentation, we start by describing how to model the distribution of single species using the probit model. For each $species_j$, we link the occurrence of $species_j$ at observation site $l_i$ with a random variable $r_{i,j}$ where the probability that $species_j$ was detected at observation site $l_i$ is equal to the probability that $r_{i,j} > 0$, i.e.

$$\Pr(b_{i,j} = 1|l_i) = \Pr(r_{i,j} > 0) \qquad (1)$$

Here, $r_{i,j}$ follows a normal distribution $N(\mu_{i,j}, \sigma)$ where $\mu_{i,j}$ is a function of $l_i$ and $\sigma$ is fixed to be 1. According to the definition of normal distribution, a positive $\mu_{i,j}$ implies that the $species_j$ is more likely to be present than absent at site $l_i$ and a negative $\mu_{i,j}$ implies the opposite. Therefore, we can model the distribution of each species by parameterizing $\mu_{i,j}$.

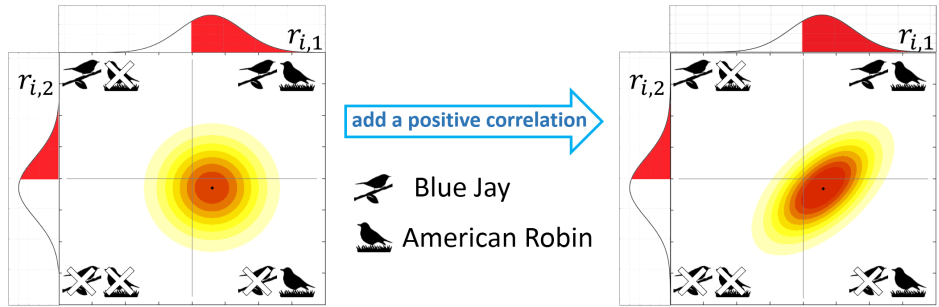

Figure 2: The left graph depicts the independent joint distribution of two species where the color from light yellow to red (better view in color) represents the probability from low to high. The probability mass in each quadrant represents the probability of each two-species co-occurrence. For example, the first quadrant represents the probability that two species occur together. The right graph is derived from the left one by adding a positive correlation between two species. Here, we plot the marginal distribution for each species on the upper side and left side of each graph. One can see, though we change the correlation between two species, the distribution of each species, unconditional of the other, remains the same.

A general approach to model the joint distribution of multiple species is to simply join the distribution of each species assuming each species is independent. For the ease of presentation, we call this kind of joint distribution "independent joint distribution".

The left graph in the picture above (Fig.2) depicts the independent joint distribution of two species (American Robin and Blue Jay) corresponding to random variable $r_{i,1}$ and $r_{i,2}$. In the graph, the color from light yellow to red represents the probability from low to high and the probability mass in each quadrant represents the probability of each co-occurrence. For example, the probability mass in the first quadrant shows the probability that American Robin and Blue Jay are present together at the observation site $l_i$. The one-dimensional distributions on the graphs' upper side and left side are the marginal distributions for American Robin and Blue Jay respectively. The red area in each one-dimensional distribution represents the probability of the presence of each species unconditional of the other.

Since the independent joint distribution fails to model the correlation between species which widely exists in the real world, we upgrade the probit model by applying multivariate normal distribution over the $n$-dimensional random variables $r_i = (r_{i,1}, ..., r_{i,n})$ i.e.

$$r_i \sim N_n(\mu_i, \Sigma) \qquad (2)$$

where $\mu_i = (\mu_{i,1}, ..., \mu_{i,n})^T$ and $\Sigma$ is the covariance matrix. In this way, each random variable $r_{i,j}$ still follows a normal distribution, but we can capture interspecies correlation by parameterizing the covariance matrix $\Sigma$.

As shown in the right graph of Fig 2, we change the covariance between random variable $r_{i,1}$ and $r_{i,2}$ from 0 to a positive number $\rho$, then the joint distribution changes significantly. For example, the probability mass in the first quadrant becomes larger, which means these two species are more likely to be present together. Although we affect the joint distribution of two species by changing the covariance between $r_{i,1}$ and $r_{i,2}$, the marginal distribution of each random variable does not change. This means the probability of the presence of each species, unconditional of the other, is unaffected by the covariance. This property ensures that our model can maintain the predictive capability derived from learning habitat preferences of each species. Meanwhile, it can outperform the independent version when the species distributions are correlated. In addition, if we restrict the variance of each species to be 1, the matrix $\Sigma$ becomes a correlation matrix, a convenient and intuitive parameterization, from which we can further have a quantitative understanding about the species-species interactions, which are only understood qualitatively among ecologists.

## 2.2 Deep Multi-Species Embedding

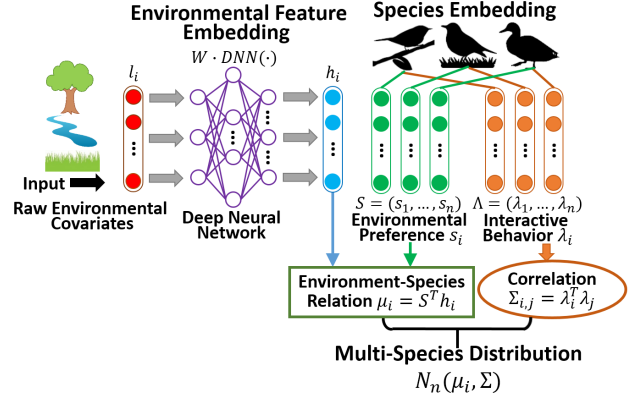

Figure 3: The intuitive visualization of DMSE framework.

In order to estimate the parameters $\mu$ and $\Sigma$, we need to first model the species-environment relationship as well as the correlation between species. To achieve this, we first embed each $species_i$ with two vectors $s_i \in R^{d_1}$ and $\lambda_i \in R^{d_2}$ representing its environmental preference and interactive behavior respectively. Here $d_1$, $d_2$ are the dimensionality of these two vector spaces which are manually set[1]. We choose to model these two characteristics separately instead of embedding to the same vector, because it is not uncommon for groups of species to share similar environment-distribution relationships, but have very different inter-species associations. Thus, by modeling these characteristics separately, DMSE can capitalize the shared environment-distribution relationships without biasing the inter-species correlation estimates. Moreover, because the environmental features used in the model describe habitat characteristics at a much coarser spatial resolution than that of the inter-species interactions, this model formulation can be seen as multi-scale approach that shares information at coarse scales while simultaneously allowing fine-scale variation between species.

When it comes to the environmental features, we apply a deep neural network and a projection matrix to embed the low-dimensional raw environmental data into the same $d_1$-dimensional feature space as the vectors $s_i$. For each observation $(b_i, l_i)$,

$$l_i \xrightarrow{embed} h_i : h_i = W \cdot DNN(l_i), \qquad (3)$$

here $DNN(\cdot)$, a function mapping from $R^m$ to $R^{n_{output}}$, represents a deep neural network[2]. $W$, a $d_1$-by-$n_{output}$ projection matrix, is used for modulating the data range and mapping the DNN's output layer to the same high-dimensional feature space with $s_i$, and $n_{output}$ is the dimension of output layer. We embed the environmental features via the deep neural network to enhance the predictive power of our model. We will include more discussion about the performance of the neural network in the experimental section.

---

[1]In our experiments, we set $d_1 = d_2 = 100$. Because embedding methods are able to take advantage of high-dimensional representations, it is advantageous to set this parameters to be high, though the methods are not sensitive to the exact values.

[2]In our experiment, we empirically found that a 3-hidden-layer fully connected neural network using tanh as the activation function worked the best. The number of neurons in each hidden layer was 256, 256, 64.

In order to simplify the presentation, we concatenate the vectors $s_i$ and $\lambda_i$ as the columns into two matrices.

$$S = (s_1, s_2, ..., s_n) \in R^{d_1 \times n},$$
$$\Lambda = (\lambda_1, \lambda_2, ..., \lambda_n) \in R^{d_2 \times n} \quad (4)$$

Using the notations in equation(2), (3) and (4), we can formulate our DMSE model as follows,

$$\Pr(b_{i,j} = 1|l_i) = \Pr(r_{i,j} > 0), \quad r_i \sim N_n(\mu_i, \Sigma), \quad (5)$$
$$\text{where } \mu_i = S^T h_i = S^T(W \cdot DNN(l_i)) \text{ and } \Sigma = \Lambda^T \Lambda.$$

Here $\mu_{i,j} = s_j^T h_i$ scores the habitat suitability of $species_j$ at observation site $l_i$ and $\Sigma_{i,j} = \lambda_i^T \lambda_j$ represents the correlation between $species_i$ and $species_j$. According to the definition of multivariate normal distribution, we derive that

$$\Pr(b_i|l_i) = \int_{L_1}^{R_1} ... \int_{L_n}^{R_n} f(x) dx_1 ... dx_n \quad (6)$$
$$\text{where } f(x) = \frac{1}{\sqrt{(2\pi)^n |\Sigma|}} \exp(-\frac{1}{2}(x - \mu)^T \Sigma^{-1}(x - \mu)),$$
$$\text{and } L_j = \begin{cases} 0 & \text{if } b_{i,j} = 1 \\ -\infty & \text{if } b_{i,j} = 0 \end{cases}, R_j = \begin{cases} +\infty & \text{if } b_{i,j} = 1 \\ 0 & \text{if } b_{i,j} = 0 \end{cases}$$

Finally, we train our model by maximizing the log-likelihood on the observation data. The parameters that should be trained are the matrix $S, \Lambda, W$ and the parameters in the deep neural network denoted by $\theta_{DNN}$.

$$S, \Lambda, W, \theta_{DNN} = \text{argmax}_{S,\Lambda,W,\theta_{DNN}} \sum_{i=1}^{N} \log \Pr(b_i|l_i) \quad (7)$$

## 2.3 Training and Testing

We use the stochastic gradient descent algorithm as proposed in [6, 2] to optimize the log-likelihood function in equation (7). In order to train and test our DMSE model, we need to be able to compute the integration in equation (6) and its derivatives with respect to each parameter.

For the integration part, we use an adaptive algorithm proposed in [10], implemented using function $mvn$ in python (scipy.stats) which can calculate the cumulative distribution function (CDF) on multivariate normal distribution with a high accuracy (relative error $< 1e - 6$).

To compute the derivative of $\Pr(b_i|l_i)$, one key observation is that if we can compute the derivative of $\Pr(b_i|l_i)$ with respect to $\mu$ and $\Sigma$, we can easily obtain other derivatives we want by simply applying the chain rule. Since the multivariate normal distribution is uniformly continuous, we first transform the

derivative of the integration into the integration of the derivative of density function as follows.

$$\frac{\partial \log \Pr(b_i|l_i)}{\partial \mu} = \frac{1}{\Pr(b_i|l_i)} \int_{L_1}^{R_1} ... \int_{L_n}^{R_n} \frac{\partial f(x)}{\partial \mu} dx_1...dx_n$$
$$\frac{\partial \log \Pr(b_i|l_i)}{\partial \Sigma} = \frac{1}{\Pr(b_i|l_i)} \int_{L_1}^{R_1} ... \int_{L_n}^{R_n} \frac{\partial f(x)}{\partial \Sigma} dx_1...dx_n$$
(8)

Using the definition of multivariate normal distribution, we derive the following equations:

$$\frac{\partial f(x)}{\partial \mu} = f(x) \cdot F(\Sigma, \mu, x), \frac{\partial f(x)}{\partial \Sigma} = f(x) \cdot G(\Sigma, \mu, x)$$
where $F(\Sigma, \mu, x) = \Sigma^{-1}(x - \mu)$,
$$G(\Sigma, \mu, x) = -\frac{1}{2}(\Sigma^{-1} - \Sigma^{-1}(x - \mu)(x - \mu)^T \Sigma^{-1})$$
(9)

According to equation (6), we know that

$$\int_{L_1}^{R_1} ... \int_{L_n}^{R_n} \frac{f(x)}{\Pr(b_i|l_i)} dx_1...dx_n = 1 \tag{10}$$

Thus, we can consider $\frac{f(x)}{\Pr(b_i|l_i)}$ as the density function of a distribution over a hyper-cube $Q \subseteq R^n$ corresponding to the integration range of equation(10). Thus, we can employ the Markov Chain Monte Carlo sampling method to estimate the derivative of $\log \Pr(b_i|l_i)$ with respect to $\mu$ and $\Sigma$ as follows:

$$\frac{\partial \log \Pr(b_i|l_i)}{\partial \mu} = \int_{L_1}^{R_1} ... \int_{L_n}^{R_n} \frac{f(x)}{\Pr(b_i|l_i)} F(\Sigma, \mu, x) dx_1...dx_n$$
$$= E\Big[F(\Sigma, \mu, x)\Big]_{x \in Q} \approx \frac{1}{M} \sum_{k=1}^{M} F(\Sigma, \mu, x_k) \tag{11}$$
$$\frac{\partial \log \Pr(b_i|l_i)}{\partial \Sigma} = \int_{L_1}^{R_1} ... \int_{L_n}^{R_n} \frac{f(x)}{\Pr(b_i|l_i)} G(\Sigma, \Sigma, x) dx_1...dx_n$$
$$= E\Big[G(\Sigma, \mu, x)\Big]_{x \in Q} \approx \frac{1}{M} \sum_{k=1}^{M} G(\Sigma, \mu, x_k) \tag{12}$$

To make our model more efficient, we apply an enhancement for our model. Using the property of normal distributions, we know that $\Pr(|r_{i,j} - \mu_j| > k\Sigma_{j,j}) < \frac{e^{-k^2/2}}{k\sqrt{2\pi}}$. As the result, we can make a cut-off on $L_i$ and $R_i$ which significantly reduces our sample range and increases the convergence rate in our sampling process.

## 3 Related Works

We refer the reader to [7] for a survey of general techniques used in species distribution modeling. Modeling approaches in this area vary depending on the type of observational data and application objectives. The most commonly available observational data records only where species have been detected and identified, known as presence-only data. The authors of [25] developed the popular MaxEnt model using maximum entropy to estimate the population intensity. More recently, the connections between Poisson point processes and MaxEnt have been used to develop presence-only data models [9]. Other data collection protocols, like *eBird*, record both when species are and are not detected. These presence-absence datasets are typically modeled using a variety of statistical and machine learning methods including additive logistic regression, random forests, and boosted regression trees. Occupancy models [20] account for imperfect detection of species by explicitly modeling hierarchically linked observation and occupancy processes, resulting in stronger ecological inferences [11, 15]. Species distribution models have also been extended to capture population dynamics using cascading models [29], Brownian Bridges [14], circuit theory [22], and non-stationary predictor response relationships [8]. Recent extensions to joint species distribution models focus on modelling the unobserved environmental factors which potentially drive the correlated distribution [12], and for spatiotemporal dynamics based on Gaussian processes [30]. In machine learning literature, multi-label classification [32, 16, 33, 26, 17, 18] is also related to our work and can be applied to multi-species modeling. Most research in multi-label classification is based on ensemble of classifier chains (ECC), which is different from our approach and cannot provide direct information about the species correlation matrix. Among these previous work, [12], which also uses the latent random variables to model correlations, is most closely related to ours. However, their model can only handle a few (no more than 10) random variables to infer the unobserved factors which potentially drive the correlated distribution, and it ignores the interaction between species. In contrast, our DMSE method can handle hundreds of latent random variables for each species and can quantitatively measure the interaction among species. In the experiments section, we show that our DMSE model outperforms many aforementioned models.

Our model is also inspired by embedding methods which are widely applied to many areas, including music [3], language [1, 23], online purchase behavior [27] etc. The core idea is to learn a vector (or other structure) to represent each of the data points, so that the interaction in the vector space reflects the semantic meaning in the original data. Embedding methods have been proven to have better generalization performance and to provide a better data visualization as well. [28] presents an embedding model that assumes an exponential family of conditional distributions, similar to Generalized Linear Models [21], to link observed quantities to latent embeddings that capture the semantic relationships of interest. Our DMSE model was developed independently of [28][3]. While the probit model used in DMSE is in the exponential family, DMSE differs fundamentally from the work in [28]: The DMSE framework considers two heterogeneous contextual information feature sets (environmental features and interspecies relationships), it uses a deep neural network at the latent quantity level to extract high-level feature from environmental covariates and it couples the environmental and species embeddings into a predictive multi-species distribution model. It would be interesting to adapt the embeddings proposed in [28] and incorporate them into our DMSE setting. To our knowledge, we are the first ones to apply embedding methods with deep neural network structure to multi-species modelling.

## 4  Experiments

We work with crowd-sourced bird observation data collected from the successful citizen science project eBird [24]. One record in this dataset is often referred to as a checklist in which the bird observer reports all the species he/she detects as well as the time and geographical location of the observation site. Crossed with the National Land Cover Database for the U.S. [13], we can estimate the landscape composition of each observation site $l_i$ with 15 different land types such as the percentage of the water, forest, grass, etc. For the use of training and testing, we transform all this data into the form $(b_i, l_i)$ as described in the first paragraph of the Multi-Species Modelling section. The dataset for this experiment is formed by picking all the observation checklists from the Bird Conservation Region [5] (BCR) 13 in the last two weeks of May from the 2002-2012 which contains 39154 observations. May is a migration period for BCR 13, therefore a lot of non-native birds pass over this region, which gives us excellent opportunities to observe their habitat choice during the migration. Here we choose the top 100 most frequently observed birds as the species collection which covers $97.6\%$ of the records in our dataset. In the experiments, we use a 5-fold cross validation to validate the multiple choices of hyper-parameters as well as evaluate the stability of models and we observe no overfitting between the loss on the validation vs test set during cross-validation.

---

[3]We thank Liping Liu and David M. Blei for bringing up to our attention and discussing the Exponential Family Embedding in personal communications.

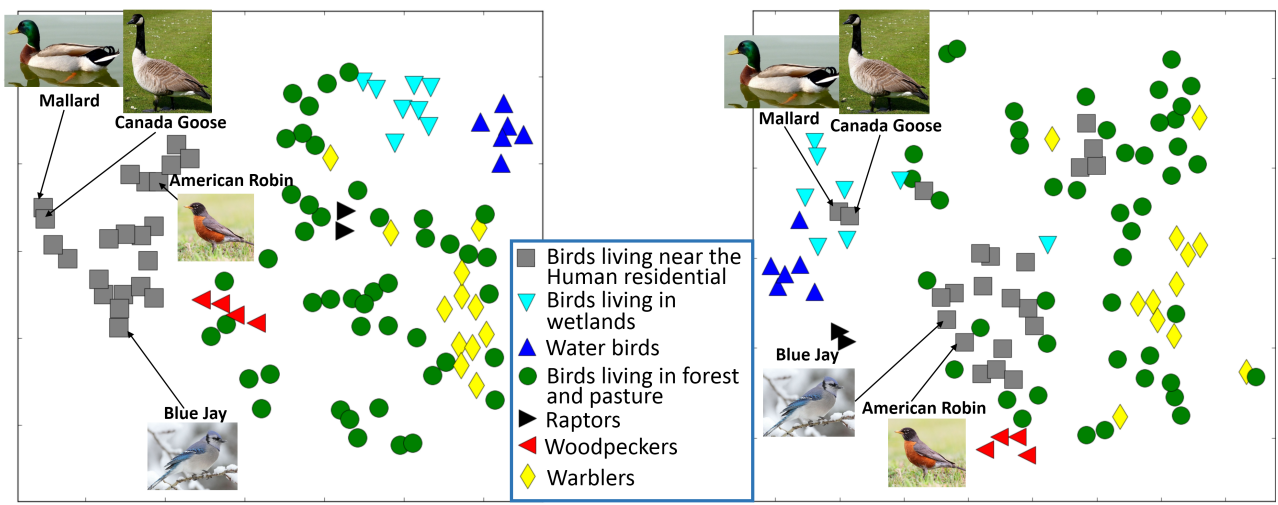

Figure 4: The left map visualizes the embeddings $s_i$ representing the environmental preference of each species and the right graph depicts the embeddings $\lambda_i$ corresponding to the correlation among species. One can see (the left map), birds of the same category cluster tightly and birds of the same breed also have a similar environmental preference. Compared with the right graph, one can find that the birds living in similar habitat have relative high correlation, but there are still some birds with high correlation that have a different environmental preference.

## 4.1 What do embeddings look like?

We start by giving a qualitative impression of the embeddings produced by our method and visualized by t-SNE algorithm [19]. Fig.4 visualizes the embeddings of environmental preference and interactive behavior ($s_i$ and $\lambda_i$) of each species. In the picture, we manually assign the species into four main categories according to their habitat preference[4]: **(1) Birds living near residential areas**, such as House Sparrow, Common Grackle, American Robin, Blue Jay, Mallard, Canada Goose, etc. Most of them are easy to find in the backyards, city parks, parking lots and agricultural fields. The presence ratio of these species are more than 25% of the records since they are easy for bird-watchers to find. **(2) Birds living in wetlands**, such as Swamp Sparrow, Northern Rough-winged Swallow, Killdeer, etc. that live near the water but mainly feed on insects. **(3) Water birds**, such as gulls, herons and cormorants which need a large amount of open water. **(4) Birds living in forest and pasture**, such as warblers, woodpeckers, nuthatches, thrushes, hawks, etc. These kinds of birds always live in the forest, grassland, pasture, shrubs, or near forest edges. These are the four categories that do not overlap with each other.

---

[4] We get the habitat preference of birds from the website www.allaboutbirds.org

One can see in the left map of Fig.4, the birds of the same category cluster tightly. For example, the birds living near the human settlements are all on the left, the birds living in wetlands and the water birds are on the right-top corner. Since the birds living in forest and pasture have a large habitat range, we further highlight three breeds in this category: the warblers, the woodpeckers and the raptors. It is interesting to note that the birds within a breed have a high similarity of habitat preference which coincides with the field observation.

When it comes to the embeddings of correlation (the right map), it can be observed that in most cases, the species living in similar places have a relative higher correlation. However, one can find some interesting cases comparing the left map and the right map. For example, although the Mallard and Canada Goose are more common to see near human habitation, the occurrence of these two birds still has a high correlation with other water birds. What is more, in the left map, we find that the locations of Blue Jay and American Robin are not very close, but from the right map, we know that they have a very high correlation which coincides with the ecological relationship as we described in the introduction section.

### 4.2 Predictive Performance of DMSE

While the visualization provides interesting qualitative insights, we now provide a quantitative evaluation of the model quality based on the predictive power. In our experiment, we analyze the performance of DMSE both on single-species modelling and multi-species modelling. Here we use two metrics to analyze the performance of each model: **(1) Area under Curve (AUC)**, i.e. the area under receiver operating characteristic (ROC) curve which is a common statistic used in predictive models. **(2) log-likelihood**, i.e. $\sum_{i=1}^{N} \log \Pr(b_i|l_i)$.

#### 4.2.1 DMSE's performance on a single-species model

We compare the single-species predictive performance of DMSE with the commonly used random forest (RF) model and SVM in terms of AUC. We implemented RFs and SVMs using python-sklearn. The number of trees in RFs was 1000, which saturated the predictive performance. The kernel of SVMs was RBFs, which perform well across a range of applications. Here we also analyze the effect of the deep neural network in the DMSE model by analyzing the performance of the DMSE model without neural network, in which we only use projection matrix $W$ to embed the environmental features. We test these four models on different species from very common to rarely seen. As shown in Fig.5, the deep neural network gives us a significant boost on the predictive power of DMSE. We expect a similar performance boost when we incorporate deep structures into other relevant models, such as the exponential family embedding model in [28]. With the help of deep neural network, our DMSE model outperforms other models.

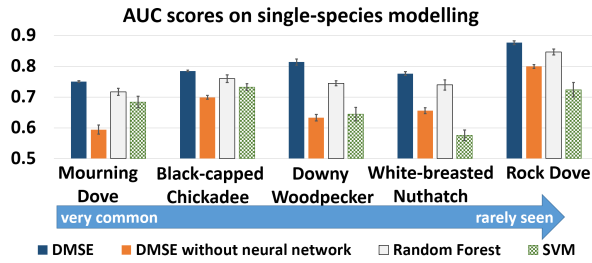

Figure 5: With the help of neural network, our single species version DMSE outperforms other models in terms of AUC.

### 4.2.2 What are the effects of correlation?

We now explore whether the correlation plays an important role in multi-species modelling. We start by comparing the performance of multi-species DMSE and the single version of DMSE on modelling two-species distribution. The single version of DMSE means we model the multi-species distribution by modelling the distribution of each species independently without their correlation. Here we use **log-likelihood** instead of **AUC** to analyze models' performance because the AUC averaged across species still values the distribution of each species separately, which does not fully reflect the benefit of modeling correlation. According to our experimental results, the multi-species DMSE outperforms the single version on all the species pairs that we have tried. Because of space limitation, we only show the performance on 3 pairs of species.

| Species Name | Species Name | correlation |
|---|---|---|
| Red-eyed Vireo | Eastern Wood Pewee | 0.607 |
| Common Grackle | Red-winged Blackbird | 0.604 |
| European Herring Gull | Great Black-backed Gull | 0.580 |
| Yellow Warbler | Common Yellowthroat | 0.567 |
| Blue Jay | American Robin | 0.535 |
| Common Grackle | American Robin | 0.510 |
| Blue Jay | Northern Cardinal | 0.504 |
| American Crow | American Robin | 0.493 |
| Common Grackle | European Starling | 0.475 |
| European Starling | Red-winged Blackbird | 0.474 |

Table 1: The list for species pairs with high correlation. The correlation here is derived from covariance matrix $\Sigma$.

As shown in Fig.6, multi-species DMSE has a substantial improvement compared with the single version of DMSE, which reflects the important role of inter-species correlation in ecology process. Furthermore, we provide Table. 1, which quantitatively measures the interaction between some species

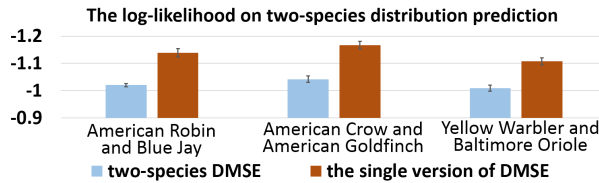

Figure 6: By modelling the correlation, the two-species DMSE outperforms the single version.

pairs with relatively high correlation.

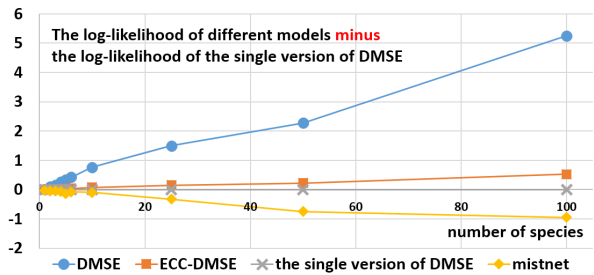

Figure 7: As the number of species becomes larger, the performance of multi-species DMSE becomes better and better compared with single species DMSE and other models. This figure shows the performance difference of all models against the single species DMSE model.

In addition, we compare our DMSE model with **(1) the single version of DMSE**, **(2) the ensemble of classifier chains (ECC) method** [17] which is a popular approach in multi-label classification (We use single version of DMSE as building blocks of the classifier chain) and **(3) MISTNET model** proposed in [12], a recent extension of species distribution model that models unobserved environmental factors which potentially drive the correlated distribution of multiple species. Here we compare all models against the single version of DMSE.

In Fig.7, as the number of species goes up, the predictive performance of our multi-species DMSE keeps improving and it outperforms other models. We believe that the ensemble of classifier chain method does not perform well mainly because of errors keep cumulating further down the classifier chain. This experiment not only highlights the importance of modelling the correlation between species, but also shows the improvement of DMSE model over previous approaches.

## 5  Conclusion

We present a novel Deep Multi-Species Embedding model that can quantitatively capture interspecies correlations of hundreds of species simultaneously, by jointly embedding vectors corresponding to multiple species as well as vectors representing environmental covariates into a common high-dimensional feature space via a deep neural network. Our DMSE model significantly outperforms existing models on multi-species distribution modelling. Additionally, we demonstrate the benefit of using a deep neural network for feature extraction and show how they improve the predictive performance of species distribution modelling. The ability to visualize the learned embeddings is also a key feature for easy interpretability and open-ended exploratory data analysis. Furthermore, the embedding models described in the paper can easily be adapted and extended to include further information (e.g., spatio-temporal information), providing many directions for future work.